%% file: paper.tex
\documentclass{article}

\PassOptionsToPackage{numbers,compress}{natbib}
\usepackage[main,final]{neurips_2026}

\usepackage[utf8]{inputenc}
\usepackage[T1]{fontenc}
\usepackage{hyperref}
\usepackage{url}
\usepackage{booktabs}
\usepackage{amsfonts}
\usepackage{amsmath}
\usepackage{amssymb}
\usepackage{amsthm}
\usepackage{mathtools}
\usepackage{nicefrac}
\usepackage{microtype}
\usepackage{xcolor}
\usepackage{graphicx}
\usepackage{multirow}
\usepackage{enumitem}
\usepackage{tabularx}
\usepackage{xspace}
\usepackage{array}
\usepackage{needspace}
\usepackage{placeins}

\newtheorem{theorem}{Theorem}
\newtheorem{proposition}{Proposition}

\newcommand{\methodname}{\textup{\textsc{BASIS}}\xspace}
\newcommand{\methodlong}{Bayesian Axis Selection via Interactive Signals\xspace}
\newcommand{\Z}{\mathcal{Z}}
\newcommand{\A}{\mathcal{A}}
\newcommand{\Q}{\mathcal{Q}}
\newcommand{\R}{\mathbb{R}}
\newcommand{\E}{\mathbb{E}}

\newcommand{\argmax}{\operatorname*{argmax}}

\title{Bayesian Test-Time Inference of Task-Aligned Similarity from Weak Interactive Feedback}

\author{%
Weida Liang \\
National University of Singapore \\
\texttt{weida\_liang@u.nus.edu}
\And
Kenji Kawaguchi \\
National University of Singapore \\
\texttt{kenji@comp.nus.edu.sg}
}

\begin{document}

\maketitle

\input{sections/00_abstract}
\input{sections/01_introduction}
\input{sections/02_related_work}
\input{sections/03_problem_formulation}
\input{sections/04_method}
\input{sections/05_theory}
\input{sections/06_experimental_setup}
\input{sections/07_results}
\input{sections/08_analysis}
\input{sections/09_limitations}
\input{sections/10_conclusion}

\begin{ack}
This work received no external funding. The authors declare no competing interests.
\end{ack}

\clearpage
{\small
\bibliographystyle{plainnat}
\bibliography{references}
}

\appendix
\input{sections/appendix}

\clearpage
\input{checklist.tex}

\end{document}

%% file: sections/00_abstract.tex
\begin{abstract}
Language model systems repeatedly make test-time decisions by comparing similarity: they retrieve exemplars, rank candidates, and revise responses. Yet similarity is not unique: lexical, semantic, structural, and stylistic resemblance can point to different candidates, and the task-aligned notion can change across instances. We introduce \methodname (Bayesian Axis Selection via Interactive Signals), a training-free Bayesian method that uses weak test-time feedback to infer which criterion in a supplied, task-specific library governs the current decision. It maintains a joint posterior over candidate targets and criteria, updates it from scalar scores, pairwise preferences, or verifier outputs, and selects probes by expected information gain. In a finite noisy model, we characterize when the target--criterion pair is recoverable and show why fixed-criterion and fixed-mixture rules can fail under criterion shift. Across concept recovery, reasoning exemplar selection, and response refinement, \methodname improves six-probe concept-recovery Success@1 from 56.0\% to 68.1\%, reasoning accuracy from 72.3\% to 73.1\%, and refinement win rate from 61.4\% to 63.8\% over tuned static mixtures. It also exceeds compute-matched best-of-$N$ selection on reasoning (72.1\%) and refinement (61.2\%). Gains are largest under criterion mismatch and with larger candidate pools, and matched controls indicate that adaptive criterion weighting explains much of the benefit, with joint target--criterion inference adding further value. \methodname selects only among supplied criteria and does not discover new ones.
\end{abstract}

%% file: sections/01_introduction.tex
\section{Introduction}\label{sec:intro}

Language models rank passages, demonstrations, and response drafts using similarity
\citep{brown2020language,lewis2020retrieval,rubin2022learning}. Yet relevance depends on the
instance: a useful math exemplar may share the operation sequence rather than the topic,
while a preferred revision may match the requested tone rather than the subject matter.
A fixed embedding or mixture score may select a plausible but task-misaligned candidate.

At test time, systems may receive scalar, pairwise, or verifier feedback. Prior work uses
such signals for reranking, query rewriting, self-correction, and verifier-guided selection
\citep{madaan2023self,shinn2023reflexion,gou2024critic,asai2024selfrag,mao2024rafe,cobbe2021training,setlur2025rewarding}.
But feedback can show which candidate works without revealing why: a verifier score may
reward structural fit, answer consistency, or style, and the useful criterion can vary by
instance. Search and stronger rerankers improve selection under a chosen scoring rule,
but do not by themselves infer when that rule should change \citep{snell2024scaling,brown2024large}.

We study this inference within a finite, task-specific criterion library supplied in advance.
\methodname (Bayesian Axis Selection via Interactive Signals) is training-free: it maintains a
joint posterior over candidate targets and criteria, updates it from scalar, pairwise, or
verifier feedback, and selects probes by expected information gain. Joint inference retains
target--criterion dependence, while the criterion marginal makes the selection process
inspectable.

After six probes, concept-recovery Success@1 is 68.1\% for \methodname versus 56.0\% for a
tuned mixture. Reasoning accuracy is 73.1\% versus 72.3\%, and refinement win rate is
63.8\% versus 61.4\%. Under distractors, reasoning drops by 4.9 points for \methodname and
6.5 for the tuned mixture. We evaluate one model stack and inference within a supplied criterion library.

\Needspace{7\baselineskip}
Our contributions are:
\begin{enumerate}[leftmargin=1.2em]
    \item We formulate feedback-driven selection as joint inference over a candidate and a
    finite library of relevance criteria, and introduce \methodname for this task.
    \item We characterize recovery under separability and probe coverage, and identify settings
    in which a fixed criterion or mixture cannot track instance-dependent relevance.
    \item We evaluate concept recovery, reasoning exemplar selection, and response refinement,
    using matched controls to separate adaptive weighting from joint target--criterion inference.
\end{enumerate}

%% file: sections/02_related_work.tex
\section{Related Work}\label{sec:related}

\paragraph{Retrieval and example selection.}
Few-shot prompting and RAG select demonstrations or passages \citep{brown2020language,lewis2020retrieval}.
Prompt retrieval, Bayesian example selection, and structure-aware retrieval improve ranking
\citep{rubin2022learning,wang2024bayesian,lin2025reasoning}. \methodname instead infers which
available criterion is relevant to each query.

\paragraph{Feedback-driven test-time improvement.}
Self-Refine, Reflexion, CRITIC, Self-RAG, RaFe, and verifier-based methods use feedback for
revision, retrieval, or candidate scoring \citep{madaan2023self,shinn2023reflexion,gou2024critic,asai2024selfrag,mao2024rafe,cobbe2021training,setlur2025rewarding}.
\methodname uses feedback to infer the criterion that best explains each signal.

\paragraph{Conditional retrieval.}
C-STS, FollowIR, and perspective-aware retrieval condition similarity on an explicit aspect,
instruction, or viewpoint \citep{deshpande2023csts,weller2025followir,zhao2024beyond}.
These methods supply a criterion. \methodname infers which one in a finite library best
accounts for weak feedback.

\paragraph{Bayesian inference and active learning.}
Bayesian accounts of in-context learning view test-time behavior as posterior inference over
latent tasks \citep{ahuja2024bayesianprism}. Active learning selects queries to reduce
uncertainty \citep{mackay1992information,settles2009active}. Information Pursuit selects
interpretable queries by information gain for supervised prediction
\citep{chattopadhyay2022interpretable}. \methodname uses informative queries for training-free
inference over a joint candidate--criterion state, then ranks candidates in the pool.

\paragraph{Reranking and test-time compute.}
Richer rerankers and repeated verifier-guided sampling improve selection under fixed scores
\citep{snell2024scaling,brown2024large}. We compare these alternatives in settings where the
criterion may shift across instances.

%% file: sections/03_problem_formulation.tex
\section{Problem Formulation}\label{sec:problem}

Let $\Z$ be a finite candidate space and $\A$ a finite, task-specific library of relevance
criteria, such as semantic similarity, entity overlap, reasoning structure, or style. We call
each criterion an \emph{axis} in the notation below, following the name BASIS. The terms refer
to the same latent variable $a\in\A$. The unknown pair $(z^\star,a^\star)$ identifies a useful
candidate and the criterion governing its relevance. Our formulation assumes that a useful
criterion is represented in $\A$.

At interaction round $t$, the learner chooses a probe $q_t \in \Q$ and observes weak
feedback $y_t$. A probe can be a single candidate presented for scoring, a pair of
candidates presented for comparison, or a structured question designed to reveal one aspect
of the target. Feedback is generated from a criterion-conditioned model,
\begin{equation}
    y_t \sim p(y \mid q_t, z^\star, a^\star).
\end{equation}
The learner accumulates data $\mathcal{D}_t = \{(q_i, y_i)\}_{i=1}^t$ and maintains the joint
posterior
\begin{equation}
    p_t(z,a) = p(z,a \mid \mathcal{D}_t).
\end{equation}

The central ambiguity is that one feedback signal can often be explained by multiple
target--criterion pairs. A high score may indicate topical overlap under one criterion,
structural compatibility under another, or stylistic affinity under a third. Without modeling
$a$, a learner can assign high confidence to a wrong target supported only by the wrong notion
of similarity.

The resulting decision rule should therefore depend on the posterior over criteria. For an input
$x$ and candidate $z$, our dynamic utility is
\begin{equation}
    U_t(x,z) = \sum_{a \in \A} p_t(a) \, s_a(x,z).
\end{equation}
A static mixture replaces $p_t(a)$ with fixed weights $w_a$ and therefore cannot adapt to the
current instance or feedback history.

Our objective is to use as few rounds of feedback as possible to recover $z^\star$,
identify $a^\star$, or choose the best downstream action. In retrieval, this means selecting
exemplars with maximal task utility. In refinement, it means choosing a response aligned with
latent user preference. In concept recovery, it means localizing the hidden target among
confusable distractors. Empirically, we evaluate target recovery, downstream utility, and
whether dynamic posterior weighting can be matched by fixed mixtures, stronger rerankers,
or more brute-force search under the same signals.

%% file: sections/04_method.tex
\section{Method}\label{sec:method}

We call our framework \methodname~(\methodlong). At test time, it uses a frozen representation
backbone and weak interaction to infer both \emph{which target is useful} and \emph{which
criterion matters for the current instance}. The reasoning library spans semantic, lexical,
entity, structural, and answer-consistency criteria. Refinement uses six preference criteria.

\methodname has three moving parts: a criterion-conditioned scoring library, a Bayesian
posterior update, and an information-seeking probe policy. The scoring library defines
plausible explanations of feedback. The posterior decides which explanation is most
credible. The probe policy chooses the next interaction that will reduce uncertainty most.

\subsection{Criterion-conditioned similarity}

For every probe $q$ and candidate $z$, we obtain a base representation $h(q), h(z) \in
\R^d$ from a frozen encoder. Each criterion $a \in \A$ defines a projection or scoring rule
$P_a$, giving the criterion-conditioned similarity
\begin{equation}
    s_a(q,z) = \cos\bigl(P_a h(q), P_a h(z)\bigr).
\end{equation}
In practice, $P_a$ is a deterministic feature transform or lightweight scorer. The reusable
criteria are semantic, lexical, entity, structural, style, and answer consistency.

The six criteria form a reusable menu rather than a universal inventory. Each task uses a
relevant subset. Performance depends on the supplied library containing a useful criterion.
Section~\ref{sec:analysis} measures the effect of weakening this assumption.

\subsection{Bayesian update from weak feedback}

Starting from a prior $p_0(z,a)$, \methodname updates the joint posterior after every
feedback observation:
\begin{equation}
    p_t(z,a)
    \propto
    p(y_t \mid q_t, z, a)\, p_{t-1}(z,a).
\end{equation}
This produces two useful marginals,
\begin{equation}
    p_t(z) = \sum_{a \in \A} p_t(z,a),
    \qquad
    p_t(a) = \sum_{z \in \Z} p_t(z,a).
\end{equation}
The target marginal supports downstream selection. The criterion marginal indicates which
notion of relevance best explains the observed feedback.

\paragraph{Scalar scores.}
For scalar feedback, we use
\begin{equation}
    p(y_t \mid q_t, z, a)
    =
    \mathcal{N}\bigl(y_t; s_a(q_t,z), \sigma_a^2\bigr).
\end{equation}

\paragraph{Pairwise preferences.}
For pairwise feedback, we use a Bradley--Terry preference model,
\begin{equation}
    p(q_i \succ q_j \mid z,a)
    =
    \frac{\exp\bigl(\beta s_a(q_i,z)\bigr)}{\exp\bigl(\beta s_a(q_i,z)\bigr)+\exp\bigl(\beta s_a(q_j,z)\bigr)}.
\end{equation}

\paragraph{Verifier signals.}
For pass/fail or verifier outputs $y_t \in \{0,1\}$, we use
\begin{equation}
    p(y_t=1 \mid q_t,z,a) = \sigma\!\left(\gamma s_a(q_t,z) + b_a\right),
\end{equation}
where $\sigma(\cdot)$ is the logistic sigmoid.

Likelihoods must be sampleable and evaluable. Mutual information need not be closed form.
Scalar, pairwise, and verifier feedback use Gaussian, Bradley--Terry, and logistic models.

\subsection{Active probe selection}

At round $t$, \methodname selects the probe that maximizes expected entropy reduction:
\begin{equation}
    q_{t+1} = \argmax_{q \in \Q}
    \Bigl[
    H\bigl(p_t(z,a)\bigr)
    - \E_{y \sim p(y \mid q,\mathcal{D}_t)}
    H\bigl(p_{t+1}(z,a \mid q,y)\bigr)
    \Bigr].
\end{equation}
This quantity is exactly the mutual information $I((z,a); y \mid q, \mathcal{D}_t)$. The
greedy probe is therefore the natural one-step choice for this uncertainty-reduction
objective.

The default uses the top $K$ candidates under $p_t(z)$ and $M$ Monte Carlo feedback samples.
An optional criterion-entropy surrogate is faster but discards information about the joint state.
The results are reported in Section~\ref{sec:analysis}.

\subsection{Decision rule and complexity}

For target recovery, we return the MAP estimate
\begin{equation}
    \hat{z} = \argmax_{z \in \Z} p_T(z).
\end{equation}
For downstream retrieval or refinement, we rank candidates by $U_T(x,z)$.

With cached criterion scores, posterior construction and updating cost
$\mathcal{O}(N|\A|)$ per interaction round. Evaluating a fixed set $\Q_{\mathrm{eval}}$ of
candidate probes with top-$K$ pruning and $M$ Monte Carlo feedback samples costs
$\mathcal{O}(|\Q_{\mathrm{eval}}|K|\A|M)$. Thus, posterior updates grow linearly with the full
candidate count, while probe selection depends on the retained candidates and evaluated-probe
set. These costs exclude candidate generation and query-dependent criterion scoring.

%% file: sections/05_theory.tex
\section{Theoretical Analysis}\label{sec:theory}

We study target--criterion recovery and fixed-rule failures in a finite model. Guarantees require
separability and probe coverage and need not extend to general language-model feedback.

\subsection{Assumptions}

\paragraph{Assumption 1: finite spaces.}
The candidate space $\Z$ and criterion library $\A$ are finite, and the score functions $s_a(q,z)$ are normalized to $[-1,1]$.

\paragraph{Assumption 2: noisy scalar feedback.}
For the true pair $(z^\star,a^\star)$,
\begin{equation}
    y_t = s_{a^\star}(q_t,z^\star) + \epsilon_t,
    \qquad
    \epsilon_t \sim \mathcal{N}(0,\sigma^2).
\end{equation}
The noise is independent across rounds, and each probe is selected using only past observations.

\paragraph{Assumption 3: separability.}
For every incorrect pair $(z,a) \neq (z^\star,a^\star)$, there exists at least one probe $q \in \Q$ such that
\begin{equation}
    \bigl|s_{a^\star}(q,z^\star) - s_a(q,z)\bigr| \geq \Delta
\end{equation}
for some margin $\Delta > 0$.

\paragraph{Assumption 4: probe coverage.}
The query policy selects separating probes often enough that each incorrect pair is confronted by at least $m_T$ informative probes by round $T$.

\subsection{Posterior contraction}

\begin{theorem}[Posterior contraction]\label{thm:contraction}
Under Assumptions 1--4 and a uniform prior over $\Z\times\A$, there are constants $c,C>0$ such that, for any $\delta\in(0,1)$ satisfying $m_T\Delta^2/\sigma^2\geq C\log(|\Z||\A|/\delta)$,
\begin{equation}
    \Pr\!\left[
    1 - p_T(z^\star,a^\star)
    \leq
    |\Z|\,|\A|\,
    \exp\!\left(-c\, m_T \frac{\Delta^2}{\sigma^2}\right)
    \right]
    \geq 1-\delta,
\end{equation}
for all sufficiently large $T$.
\end{theorem}

Appendix~\ref{app:proofs} gives the derivation. The guarantee is conditional on separability
and sufficient probe coverage under the stated Gaussian feedback model.

\subsection{Why criterion inference matters}

\begin{proposition}[Fixed-criterion non-identifiability]\label{prop:inconsistency}
Suppose feedback is generated by $(z^\star,a^\star)$, but a learner restricts inference to a
fixed criterion $\bar{a}\neq a^\star$. If a wrong candidate $z^\dagger\neq z^\star$ satisfies
\begin{equation}
    p(y\mid q,z^\dagger,\bar{a})
    =p(y\mid q,z^\star,a^\star)
    \qquad \text{for all }q\in\Q\text{ and }y,
\end{equation}
then Eq.~(15) makes the wrong candidate observationally equivalent to the true pair.
\end{proposition}

The exact observational equivalence in Eq.~(15) also prevents joint inference from
distinguishing the two pairs if both are included in its support. Joint recovery instead
requires the separability and probe-coverage conditions of Theorem~\ref{thm:contraction}.
Without these conditions, a misspecified fixed criterion can leave a wrong target
observationally plausible even with unlimited feedback.

\subsection{Why fixed mixtures can fail}

\begin{proposition}[Static mixtures need not match adaptive criterion inference]\label{prop:mixture}
Suppose the task-relevant candidate is $z^+$ on both instances, but the relevant criterion changes. For two criteria $a_1 \neq a_2$, instances $x_1,x_2$, and candidates $z^+,z^-$, assume
\begin{equation}
    s_{a_1}(x_1,z^+) - s_{a_1}(x_1,z^-) = \Delta,
    \qquad
    s_{a_2}(x_1,z^+) - s_{a_2}(x_1,z^-) = -\Delta,
\end{equation}
while on $x_2$ the differences are reversed:
\begin{equation}
    s_{a_1}(x_2,z^+) - s_{a_1}(x_2,z^-) = -\Delta,
    \qquad
    s_{a_2}(x_2,z^+) - s_{a_2}(x_2,z^-) = \Delta,
\end{equation}
and let all remaining criteria contribute zero to these comparisons. Then no fixed mixture with weights shared across instances
\begin{equation}
    U_{\mathrm{mix}}(x,z;w) = \sum_{a \in \A} w_a s_a(x,z)
\end{equation}
can rank $z^+$ above $z^-$ on both instances: the score differences are $\Delta(w_{a_1}-w_{a_2})$ and $-\Delta(w_{a_1}-w_{a_2})$. An adaptive rule can rank $z^+$ first on both by concentrating on $a_1$ for $x_1$ and on $a_2$ for $x_2$.
\end{proposition}

This proposition isolates the role of criterion shift: with the same score library, a shared set of weights cannot satisfy both instance-specific rankings.

\subsection{Information-seeking probes}

\begin{proposition}[One-step information gain]\label{prop:ig}
For any probe $q$, the expected entropy reduction equals
\begin{equation}
    H\bigl(p_t(z,a)\bigr)
    -
    \E_y H\bigl(p_{t+1}(z,a \mid q,y)\bigr)
    =
    I\bigl((z,a); y \mid q, \mathcal{D}_t\bigr).
\end{equation}
\end{proposition}

Thus, the greedy probe selected by \methodname is the most informative one-step measurement for reducing uncertainty about the latent target--criterion pair.

%% file: sections/06_experimental_setup.tex
\section{Experimental Setup}\label{sec:experiments}

All methods share candidate pools, backbone, prompts, and feedback or verifier access whenever
applicable. Appendix~\ref{app:exp-details} provides construction and implementation details.

\paragraph{Tasks.}
For controlled concept recovery, we run 500 episodes with 50 candidates and up to 12 probes.
The hidden feedback criterion is sampled from five predefined criteria, and distractors create
clean and mismatch conditions. We report Success@1 after four and six probes and query
efficiency relative to the 12-probe final recovery rate. For reasoning exemplar selection, we
evaluate GSM8K, StrategyQA, and three BIG-Bench Hard tasks. Each query starts with 64 support
examples and an initial dense
shortlist of 32. Interactive methods receive at most three feedback rounds and select four
exemplars. The $N=128$ and $N=256$ conditions add candidates from the support bank, increasing
both pool size and distractor heterogeneity. For BBH, candidates come from other labeled
benchmark items, excluding the current query and items that share its underlying scenario.
All methods receive identical candidate pools, and the query's gold answer is used only for
final evaluation. For response refinement, we use 1{,}000 prompts
each from Anthropic HH, OpenAssistant, and UltraFeedback, with eight candidate responses and
up to three scalar or pairwise feedback rounds. The mismatch split adds topic-matched
responses that conflict with the preference criterion.

\paragraph{Criteria and feedback.}
Reasoning uses semantic, lexical, entity, structural, and answer-consistency scores.
Refinement uses six preference criteria. Scores are normalized within each pool, cached before
interaction, and shared with static-mixture baselines. Feedback combines scalar scores,
pairwise preferences, verifier deltas, and answer checks. Each feedback channel has an
explicit likelihood.

\paragraph{Comparisons.}
We compare retrieval and reranking, static mixtures, mean-field and criterion-marginal controls,
and a history-aware reranker. Best-of-$N$ uses four reasoning and six refinement calls, at least
BASIS's three feedback calls. The oracle is evaluation-only. Appendix~\ref{app:exp-details}
details the controls and each method's information access.

\paragraph{Metrics and implementation.}
Primary metrics are target recovery, reasoning accuracy, and preference win rate. Criterion
identification, criterion match, gap closed, and mismatch degradation are diagnostic. We
report equal-weight task macro-averages where stated. Results average five seeds. Confidence
intervals use 10{,}000 paired bootstrap samples, resampling test items and keeping all seed
outcomes for each item together. Unless noted otherwise, \methodname uses a uniform prior,
top-$K=16$ candidates, $M=32$ Monte Carlo feedback samples, and at most three downstream
feedback rounds. End-to-end timings are reported separately from local inference.

The frozen retriever is Qwen/Qwen3-Embedding-4B, the reasoning generator is
Qwen/Qwen3-8B with thinking disabled, and the shared verifier is
mistralai/Ministral-3-14B-Reasoning-2512. All final answer generation uses greedy decoding.
Seeds affect probe Monte Carlo sampling, candidate-pool construction, and bootstrap
resampling. The reasoning results therefore describe this matched, thinking-disabled setup and
are not directly comparable to benchmark scores reported with default thinking enabled. 

%% file: sections/07_results.tex
\Needspace{4\baselineskip}
\section{Results}\label{sec:results}

We report aggregate outcomes and costs. Appendix~\ref{app:extra-results} breaks them down by task.

\paragraph{Controlled concept recovery.}
After six probes, \methodname reaches 68.1\% Success@1, compared with 56.0\% for the tuned
static mixture and 79.5\% for the oracle-criterion upper bound. Criterion identification
accuracy is 80.9\%. It reaches 90\% of its final 12-probe recovery rate after a mean of 4.6
probes, compared with 6.5 for the tuned mixture. This controlled task separates target
selection from the downstream generator and shows that interaction history can identify a
useful criterion before the full probe budget is spent.

\paragraph{Downstream performance and latency.}
Table~\ref{tab:main-cost} reports task outcomes and end-to-end time, including final answer
generation for reasoning and scoring, feedback, and selection for refinement.

\begin{table}[t]
\centering
\small
\setlength{\tabcolsep}{5pt}
\caption{Task outcomes and end-to-end latency at batch size one. Times are means with standard deviations over 100 items. The verifier controls use $N=4$ calls for reasoning and $N=6$ for refinement. BASIS uses $T=3$ feedback calls.}
\label{tab:main-cost}
\begin{tabular}{llcc}
\toprule
Task & Method & Accuracy / win rate (\%) & Time (s) \\
\midrule
Reasoning & Dense retrieval & 70.0 & $4.82\pm0.11$ \\
 & Tuned mixture & 72.3 & $5.04\pm0.12$ \\
 & Best-of-4 verifier & 72.1 & $7.91\pm0.21$ \\
 & \methodname & \textbf{73.1} & $7.32\pm0.18$ \\
Refinement & Tuned mixture & 61.4 & $6.33\pm0.15$ \\
 & Best-of-6 verifier & 61.2 & $10.84\pm0.29$ \\
 & \methodname & \textbf{63.8} & $9.08\pm0.23$ \\
\bottomrule
\end{tabular}
\end{table}

The reasoning results are equal-weight macro-averages over GSM8K, StrategyQA, and aggregate
BBH: 73.1\% for \methodname and 72.3\% for the tuned mixture. The paired item-clustered
interval for the 0.8-point reasoning difference is $[-0.9,2.5]$. The 2.4-point refinement
gain has interval $[1.3,3.5]$. \methodname closes 29.6\% of the reasoning gap and 54.5\% of
the refinement gap between the tuned mixture and oracle-criterion results.

The reasoning averages are macro-averages, so we also report paired outcomes by domain.
Table~\ref{tab:disagreement-counts} counts item--seed outcomes.

\begin{table}[t]
\centering
\small
\setlength{\tabcolsep}{4pt}
\caption{Paired correctness counts over item--seed outcomes: 2,500 each for GSM8K and StrategyQA, 3,750 for BBH, and 8,750 pooled.}
\label{tab:disagreement-counts}
\begin{tabular}{lrrrr}
\toprule
Domain & Both correct & BASIS only & Tuned only & Both wrong \\
\midrule
GSM8K & 1{,}801 & 104 & 82 & 513 \\
StrategyQA & 1{,}802 & 111 & 88 & 499 \\
BBH aggregate & 2{,}321 & 180 & 150 & 1{,}099 \\
Pooled & 5{,}924 & 395 & 320 & 2{,}111 \\
\bottomrule
\end{tabular}
\end{table}

Relative to the tuned mixture, end-to-end latency increases by 2.28 seconds (45\%) for
reasoning and 2.75 seconds (43\%) for refinement. \methodname remains 0.59 and 1.76 seconds
faster than the corresponding verifier controls. Most added time is spent acquiring feedback.
Read together, the outcome and timing results show that adaptation is most useful when the
feedback signal can resolve criterion ambiguity. For routine reasoning instances with a small
accuracy difference, a static selector remains a lower-latency option.

\paragraph{Criterion mismatch and pool enlargement.}
On clean reasoning pools, \methodname scores 73.1\%. With distractors it scores 68.2\%, a
4.9-point drop. The tuned mixture drops from 72.3\% to 65.8\% (6.5 points). After two
additional feedback rounds, \methodname at $T=5$ reaches 71.4\% on the distractor split. On
refinement, clean/mismatch win rates are 63.8/61.5 for \methodname, 61.4/57.2 for the tuned mixture, and
65.8/64.4 for the oracle-criterion upper bound, corresponding to drops of 2.3, 4.2, and 1.4
points. Thus, the smaller degradation under distractors is consistent across both task
families, while the oracle results indicate that mismatch remains consequential even with
feedback-driven selection.

\begin{figure}[t]
\centering
\includegraphics[width=0.9\linewidth]{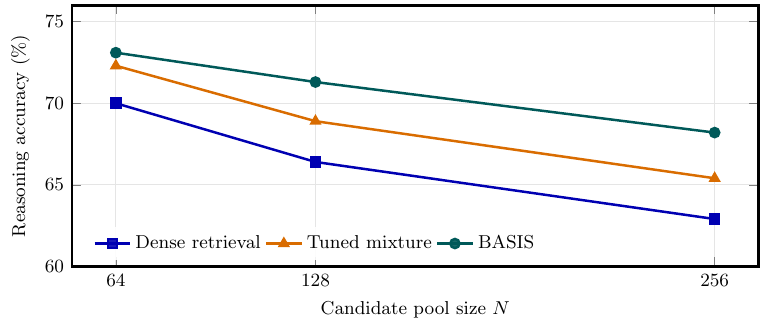}
\caption{Nested $N=64,128,256$ candidate pools grow with a heterogeneous support bank, varying both pool size and distractor composition.}
\label{fig:pool-scaling}
\end{figure}

The enlarged pools retain the original 64 task-matched candidates and add the highest-ranked
previously unused examples from a heterogeneous bank assembled from the GSM8K and StrategyQA
training splits and the BBH support manifest. The bank contains 12{,}416 unique examples. For
each query, we exclude the current query and its scenario-equivalent items, leaving at least
10{,}218 eligible examples. Candidates are unique, and every method receives the same
per-item, per-seed manifest. \methodname improves over the tuned mixture by 0.8, 2.4, and 2.8
points at $N=64,128,256$, respectively.

\paragraph{Local computational scaling.}
The cached-score benchmark measures score gathering, posterior construction and updates,
probe evaluation, and final marginalization. It excludes candidate generation, first-stage
retrieval, structural extraction, and external verifier or critic scoring. With the five
reasoning criteria, $K=16$, $M=32$, and the evaluated-probe set fixed, posterior operations
grow approximately linearly with full-pool size, while probe selection remains nearly flat.
The largest-pool timings use distinct bank examples, not replicated candidates.

\begin{table}[t]
\centering
\scriptsize
\setlength{\tabcolsep}{3pt}
\caption{Cached-score local inference costs for reasoning. Posterior storage is for the float64 joint distribution over five criteria.}
\label{tab:local-scaling}
\begin{tabular}{rccccc}
\toprule
$N$ & Gather/assembly (ms) & Update/round (ms) & Probe/round (ms) & Marginalization (ms) & Posterior (MiB) \\
\midrule
64 & 4.2 & 0.07 & 110 & 0.03 & 0.0024 \\
256 & 14.6 & 0.18 & 111 & 0.06 & 0.0098 \\
1{,}024 & 55.3 & 0.62 & 112 & 0.18 & 0.0391 \\
4{,}096 & 216.8 & 2.35 & 114 & 0.66 & 0.1563 \\
10{,}000 & 527.4 & 5.68 & 117 & 1.55 & 0.3815 \\
\bottomrule
\end{tabular}
\end{table}

At $N=10{,}000$, BASIS retains exact posterior marginalization over the full pool. We also
evaluate a two-stage setup in which dense retrieval first returns a shortlist and BASIS adapts
within it, using a separate 250-item diagnostic set described in Appendix~\ref{app:extra-results}.
First-stage retrieval is shared across selectors and excluded from local timing. BASIS timing
includes shortlist lookup, cached-score gathering, host-to-device transfer, and inference.

\begin{table*}[t]
\centering
\scriptsize
\setlength{\tabcolsep}{3pt}
\caption{Two-stage BASIS with a frozen retriever. Shortlist agreement and overlap are relative to full-pool BASIS. Exact set agreement ignores exemplar order.}
\label{tab:shortlist-scaling}
\resizebox{\textwidth}{!}{%
\begin{tabular}{rlcccccc}
\toprule
Source $N$ & Inference set & Top-1 agreement & Mean top-four overlap & Exact set agreement & Accuracy (\%) & Local time (ms) & Posterior (MiB) \\
\midrule
1{,}024 & Full pool & 100\% & 4.00/4 & 100\% & 68.4 & 393 & 0.0391 \\
1{,}024 & Top-128 & 99.0\% & 3.96/4 & 96.1\% & 68.3 & 342 & 0.0049 \\
10{,}000 & Full pool & 100\% & 4.00/4 & 100\% & 67.6 & 897 & 0.3815 \\
10{,}000 & Top-128 & 97.8\% & 3.91/4 & 91.8\% & 67.4 & 356 & 0.0049 \\
10{,}000 & Top-64 & 92.8\% & 3.72/4 & 81.4\% & 66.8 & 348 & 0.0024 \\
\bottomrule
\end{tabular}}
\end{table*}

At $N=10{,}000$, a Top-128 shortlist retains 3.91 of four exemplars on average and reaches
67.4\% accuracy versus 67.6\% for full-pool inference, while reducing local inference time
from 897 to 356 ms. These measurements concern second-stage selection, not corpus-scale retrieval.

%% file: sections/08_analysis.tex
\section{Analysis}\label{sec:analysis}

\paragraph{What joint inference adds.}
Table~\ref{tab:analysis-controls} compares the inference controls.

\begin{table}[t]
\centering
\small
\setlength{\tabcolsep}{4pt}
\caption{Matched-feedback controls. Entries are task accuracy or preference win rate in percent.}
\label{tab:analysis-controls}
\begin{tabular}{lccc}
\toprule
Method & Reasoning & Refinement & Distractor accuracy \\
\midrule
Tuned mixture & 72.3 & 61.4 & 65.8 \\
Mean-field $q(z)q(a)$ & 72.8 & 63.1 & 67.4 \\
Criterion marginal, shared probes & 72.6 & 62.7 & 67.0 \\
Criterion marginal, own probes & 72.5 & 62.4 & 66.7 \\
History-aware reranker & 72.6 & 62.6 & 67.0 \\
\methodname & 73.1 & 63.8 & 68.2 \\
\bottomrule
\end{tabular}
\end{table}

Relative to mean-field inference, the joint posterior has point-estimate gains of 0.3 points
on reasoning, 0.7 on refinement, and 0.8 on distractor accuracy. The marginal controls show a
similar pattern: the joint model adds 0.5--0.6 points on reasoning, 1.1--1.4 on refinement,
and 1.2--1.5 on distractors. Adaptive criterion weighting accounts for much of the gain,
while preserving target--criterion dependence adds a smaller benefit that varies by task.
Removing the latent criterion lowers
reasoning and refinement to 70.9 and 60.7. Disabling active probing yields 72.0 and 62.4.
The history-aware reranker receives the same three feedback observations but has no separately
optimized probe policy.

\paragraph{Probe approximation and feedback-model sensitivity.}
With $K=16$ and $M=32$, the joint-information policy selects the same probe as a $K=64$,
$M=64$ reference in 95.2\% of rounds. Local selection takes 110 ms per round, compared with
856 ms for the reference. A criterion-entropy surrogate takes 68 ms and gives 73.0/63.5 on
reasoning/refinement, compared with 73.1/63.8 for the default. Under Laplace, Student-$t_3$,
or heteroscedastic Gaussian scalar feedback while inference retains a Gaussian likelihood,
concept-recovery Success@1 after six probes is 67.5, 67.0, and 66.6, respectively, compared
with 68.1 under matched Gaussian feedback.

\begin{figure}[!htbp]
\centering
\includegraphics[width=0.9\textwidth]{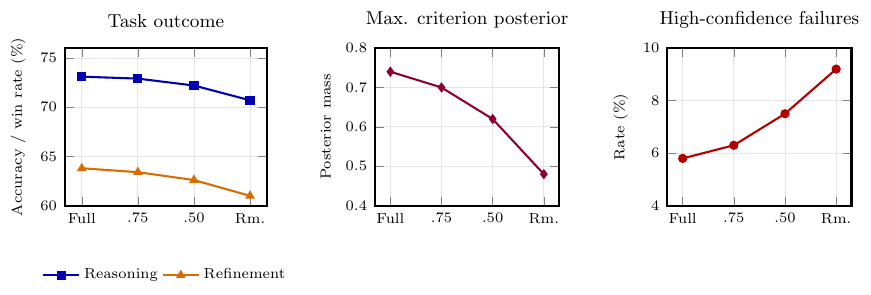}
\caption{Performance and posterior diagnostics under criterion corruption and removal. The removed-criterion condition is an oracle stress test. Maximum posterior reports the mean per-episode maximum mass over criteria.}
\label{fig:criterion-coverage}
\end{figure}

\paragraph{Coverage of the supplied criterion library.}
The dominant criterion is assigned post hoc as the criterion whose score best predicts
downstream success on held-out candidates. Removing it is an oracle stress test, not a
deployment procedure. The noisy-proxy conditions replace this score with proxy criteria at
correlations $r=0.75$ and $r=0.50$. The final condition removes it. Figure~\ref{fig:criterion-coverage} summarizes task outcomes and
posterior diagnostics as the library is corrupted or made incomplete. The maximum-criterion
posterior is computed per episode as $\max_{a\in\A}p_T(a)$, then averaged over episodes and
seeds. A high-confidence failure is an incorrect reasoning answer or non-preferred refinement
selection when posterior-predictive success is at least 0.8. This is not a calibrated guarantee.

\FloatBarrier

A fallback based on criterion entropy and posterior-predictive likelihood is thresholded on
disjoint development episodes. It achieves AUROC 0.86 and triggers on 72\% of missing-criterion
episodes versus 9\% of full-library episodes. Under criterion removal, the fallback raises
reasoning from 70.7 to 72.4 and refinement from 61.0 to 61.9, while reducing high-confidence
failures from 9.2\% to 4.8\%. It routes uncertain cases to the tuned mixture. Expanding the
reasoning/refinement libraries from 5/6 to 24 irrelevant, correlated, or redundant criteria
changes accuracy by at most 0.2 points, while local probe-selection time rises from 110 to
424 ms per round.

\paragraph{Observed selection disagreements.}
Appendix~\ref{app:case-details} gives the full sequential-fraction GSM8K trace: \methodname
selects an exemplar aligned with the updated remainder and returns \$81, while the tuned
mixture returns \$63. The contrasting BBH failure appears in Figure~\ref{fig:case-ordering}:
the tuned mixture answers hummingbird (D), but \methodname selects a direct-position exemplar
and answers cardinal (B).
\begin{figure}[!htbp]
\centering
\setlength{\fboxsep}{8pt}
\fbox{%
\begin{minipage}{0.95\textwidth}
\small
\emergencystretch=1em
\textbf{Query (BBH logical deduction).} There are five birds: an owl, a cardinal, a robin, a
hummingbird, and a crow. The hummingbird is left of the owl and right of the cardinal; the
crow is left of the robin, and the owl is left of the crow. The question asks which bird is
second from the left. The implied order is cardinal $<$ hummingbird $<$ owl $<$ crow $<$ robin,
so the answer is hummingbird (D).

\medskip
\noindent
\begin{minipage}[t]{0.47\linewidth}
\textbf{Tuned-mixture-only exemplar.} Five books are blue, purple, yellow, black, and green.
The purple book is third; yellow is left of black; green is left of purple; and blue is left of
green. The question asks which is second. It requires composing relations; the answer is green
book (E).
\end{minipage}\hfill
\begin{minipage}[t]{0.47\linewidth}
\textbf{BASIS-only exemplar.} Five birds are a blue jay, cardinal, hawk, hummingbird, and
quail. The quail is left of the blue jay; the cardinal is third; the quail is second; and the
blue jay is left of the hummingbird. The answer is quail (E), stated directly without
transitive reconstruction.
\end{minipage}

\medskip\hrule\medskip
\textbf{Selection and feedback.} The methods share three of four exemplars. The tuned mixture
answers hummingbird (D) correctly; \methodname answers cardinal (B). Verifier-score
differences for the direct-position bird and relation-composition book are $+0.286$ and
$+0.091$; the final pairwise feedback favors the direct-position example with
posterior-predictive probability 0.684. From the first feedback update to the final round,
the semantic, structural, and answer-consistency posterior masses change from
$(0.309,0.202,0.198)$ to $(0.276,0.172,0.266)$.
\end{minipage}}
\caption{BBH ordering failure. Feedback favors a direct-position exemplar, but it does not support the transitive relation needed to answer the query.}
\label{fig:case-ordering}
\end{figure}
\FloatBarrier

An HH refinement trace raises the empathy posterior from 0.17 to 0.55; when humor is absent,
BASIS favors formality. Figures~\ref{fig:case-ordering} and~\ref{fig:case-fractions} show the
reasoning queries and feedback traces.

%% file: sections/09_limitations.tex
\section{Limitations and Broader Impacts}\label{sec:limitations}

\paragraph{Criterion coverage.}
The criterion library is specified in advance. If a relevant criterion is missing or
corrupted, the posterior may favor an incorrect explanation. The evaluated fallback routes
low-fit cases to the tuned mixture, but cannot establish that the library is complete.

\paragraph{Computation and deployment scope.}
Timing starts from cached representations and criterion scores, excluding candidate
generation, first-stage retrieval, structural extraction, and one-time external scoring.
Posterior updates scale linearly with pool size. Full-pool scoring can remain costly for
criteria requiring external models.

\paragraph{Feedback and generalization.}
Most feedback comes from offline verifiers or rubric scores. Live human feedback and preference
drift remain untested. Experiments use one retriever, generator, and verifier stack, leaving
cross-model and serving-system generalization open.

\paragraph{Broader impacts.}
Task-aligned selection may improve relevance, but designers choose the criteria and feedback.
Bias or error can reinforce undesirable outputs. High-stakes use requires independent
validation and human oversight, since posterior confidence does not establish safety or fairness.

%% file: sections/10_conclusion.tex
\section{Conclusion}\label{sec:conclusion}

We study test-time candidate selection under ambiguity about which supplied criterion matters.
Across our experiments, gains over strong fixed baselines are modest on average and larger for
response refinement and criterion mismatch. Missing-criterion tests show that BASIS complements
fixed selectors but does not provide open-world criterion discovery.

%% file: sections/appendix.tex
\section{Additional Algorithmic Details}\label{app:algorithm}

\paragraph{Algorithm in words.}
Given an input $x$, a candidate pool $\Z$, and a criterion library $\A$, \methodname initializes a prior $p_0(z,a)$. At every interaction round it scores probe candidates under each criterion, estimates the mutual information objective in Section~\ref{sec:method}, chooses the most informative probe, receives weak feedback, and performs a Bayesian update. After $T$ rounds it returns a target posterior $p_T(z)$ and a criterion posterior $p_T(a)$ for retrieval, refinement, or steering.

\paragraph{Practical defaults.}
In all reported experiments we use top-$K=16$ pruning, $M=32$ Monte Carlo samples, and a maximum of three interaction rounds on downstream tasks. We found these settings sufficient to capture most of the information-gain benefit while keeping latency low.


\section{Proofs}\label{app:proofs}

\paragraph{Proof of Theorem~\ref{thm:contraction}.}
Fix an incorrect pair $h=(z,a)$ and write
\[
\delta_t=s_{a^\star}(q_t,z^\star)-s_a(q_t,z),\qquad
S_T(h)=\sum_{t=1}^T\frac{\delta_t^2}{2\sigma^2}.
\]
The log-likelihood ratio of the true pair to $h$ is
\[
\Lambda_T(h)=S_T(h)+M_T(h),\qquad
M_T(h)=\sum_{t=1}^T\frac{\delta_t\epsilon_t}{\sigma^2}.
\]
Because each probe is selected from past observations and the Gaussian noise is independent,
for any $\lambda\in(0,1)$ the process
$\exp(-\lambda M_T-\lambda^2 S_T)$ has expectation one. At $\lambda=1/4$, Markov's
inequality gives
\[
\Pr\bigl(M_T\leq-S_T/2,\ S_T\geq s_0\bigr)\leq e^{-s_0/16}.
\]
Assumption~4 implies $S_T(h)\geq s_0=m_T\Delta^2/(2\sigma^2)$. Thus, except with
probability at most $\exp(-m_T\Delta^2/(32\sigma^2))$, we have
$\Lambda_T(h)\geq m_T\Delta^2/(4\sigma^2)$. A union bound over the at most
$|\Z||\A|-1$ incorrect pairs shows that this holds for all of them with probability at least
$1-|\Z||\A|\exp(-m_T\Delta^2/(32\sigma^2))$.

Under a uniform prior, posterior odds against the true pair equal the likelihood ratios
$\exp(-\Lambda_T(h))$. On the same event, summing these odds gives
\[
1-p_T(z^\star,a^\star)
\leq |\Z||\A|\exp\!\left(-m_T\Delta^2/(4\sigma^2)\right).
\]
Taking universal constants $C=32$ and $c=1/4$ proves the theorem.

\paragraph{Proof of Proposition~\ref{prop:inconsistency}.}
By assumption, the conditional feedback distributions under $(z^\dagger,\bar{a})$ and
$(z^\star,a^\star)$ are identical for every probe. Therefore every adaptive probe policy
induces the same distribution over feedback histories under these two explanations, so no
test based on that history can distinguish them. A learner restricted to $\bar{a}$ cannot
detect that its best-fitting target may be wrong.

\paragraph{Proof of Proposition~\ref{prop:mixture}.}
For any fixed weights, the score difference between $z^+$ and $z^-$ on $x_1$ is
$\Delta(w_{a_1}-w_{a_2})$, while on $x_2$ it is $-\Delta(w_{a_1}-w_{a_2})$.
These differences cannot both be strictly positive. Instance-dependent weights concentrated
on $a_1$ for $x_1$ and $a_2$ for $x_2$ make both differences positive, proving the claim.

\paragraph{Proof of Proposition~\ref{prop:ig}.}
The identity follows from the definition of conditional mutual information:
\begin{equation}
    I\bigl((z,a);y \mid q,\mathcal{D}_t\bigr)
    =
    H\bigl(p_t(z,a)\bigr)
    -
    \E_y H\bigl(p_{t+1}(z,a\mid q,y)\bigr).
\end{equation}
Therefore, maximizing expected entropy reduction is equivalent to maximizing one-step information gain.

\section{Additional Experimental Details}\label{app:exp-details}

\begin{table}[htbp]
\centering
\scriptsize
\setlength{\tabcolsep}{3pt}
\caption{Three domains used to instantiate the same latent-criterion inference problem.}
\label{tab:task_instantiations}
\begin{tabular}{p{0.17\linewidth}p{0.20\linewidth}p{0.17\linewidth}p{0.19\linewidth}p{0.15\linewidth}}
\toprule
Task & Candidate space $\Z$ & Probe type $q_t$ & Latent criteria $\A$ & Weak feedback \\
\midrule
Concept recovery & lexical concepts with relation-based distractors & single concept or pairwise query & semantic, relation, part--whole, morphology, association & scalar score or pairwise preference \\
Exemplar selection & exemplar bank from training problems & single exemplar or exemplar pair & semantic, lexical, entity, structural, answer-consistency & verifier delta, structural critic, pairwise preference \\
Response refinement & candidate responses for one instruction & response or response pair & helpfulness, specificity, conciseness, safety, empathy, formality & preference label or rubric score \\
\bottomrule
\end{tabular}
\end{table}

\begin{table}[htbp]
\centering
\scriptsize
\setlength{\tabcolsep}{4pt}
\caption{Feedback sources used in each task. These signals are holistic weak judgments: they do not reveal the hidden criterion label, and they are available to every non-oracle method that uses external feedback.}
\label{tab:feedback_sources}
\begin{tabular}{p{0.17\linewidth}p{0.15\linewidth}p{0.27\linewidth}p{0.10\linewidth}p{0.16\linewidth}}
\toprule
Task & Feedback type & Source & Form & Shared with non-oracle baselines? \\
\midrule
Concept recovery & scalar similarity / pairwise preference & simulated from the hidden criterion with Gaussian or Bradley--Terry noise & scalar / pairwise & yes \\

Reasoning exemplar selection & verifier delta & downstream answer verifier applied to candidate exemplars or prompted traces & scalar & yes \\

Reasoning exemplar selection & structural signal & template overlap or structural critic score computed from reasoning skeletons & scalar & yes \\

Response refinement & preference feedback & dataset preference label when available, otherwise rubric-based weak supervision & pairwise / scalar & yes \\
\bottomrule
\end{tabular}
\end{table}

\paragraph{Feedback and criterion transparency.}
Except for the diagnostic oracle-criterion upper bound, no method observes the hidden feedback-generating criterion directly. Feedback is a weak scalar, binary, or pairwise judgment about a probe or candidate. In reasoning, the oracle emits holistic verifier or critic scores rather than labels such as ``structure'' or ``style.''

\paragraph{Concept recovery protocol.}
Each episode samples one hidden target, five semantically close distractors, five relation-linked distractors, five morphological distractors, and random negatives until the pool size reaches 50. We report averages over 500 episodes and five seeds. For pairwise feedback we set $\beta=10$, for scalar feedback we normalize scores to $[0,1]$.

\paragraph{Reasoning exemplar selection protocol.}
We evaluate 500 GSM8K items, 500 StrategyQA items, and 750 BBH items (250 per task). Candidate pools contain 64 exemplars, with an initial dense shortlist of 32. The final prompt includes the top four exemplars. For BBH, candidates are drawn from other labeled benchmark items, excluding the current query and items that share its underlying scenario. All methods use identical candidate pools, and the query's gold answer is used only for final evaluation. Structural similarity is computed from automatically extracted reasoning templates based on operator sequences, decomposition markers, and answer format. Exact-match or task-specific correctness is used as the end metric. For the distractor split, eight semantically close but structurally irrelevant exemplars are injected into each candidate pool.

\paragraph{Response refinement protocol.}
We sample 1{,}000 prompts per dataset, generate or retrieve eight candidate responses, and run up to three interactive rounds. Preference win rates are measured against the initial response, while attribute matching measures whether the selected response aligns with the hidden preference criterion used to generate weak feedback. The mismatch split augments each pool with topic-matched responses that are deliberately misaligned with the hidden rubric dimension.

\begin{table}[htbp]
\centering
\scriptsize
\setlength{\tabcolsep}{4pt}
\caption{Implementation of the criterion-conditioned scorers. Scores are computed once for the current candidate pool, normalized to a comparable range, and cached across interaction rounds. Feedback updates posterior weights over criteria rather than recomputing base scores.}
\label{tab:axis_implementation}
\begin{tabular}{p{0.14\linewidth}p{0.35\linewidth}p{0.16\linewidth}p{0.12\linewidth}p{0.13\linewidth}}
\toprule
Criterion & Implementation & Normalization & External model? & Updated after feedback? \\
\midrule
Semantic & cosine similarity in the frozen encoder space & pool-wise min--max to $[0,1]$ & no & no \\
Lexical & BM25 or token-overlap relevance & pool-wise min--max to $[0,1]$ & no & no \\
Entity & overlap F1 over entities, slots, or variables & native $[0,1]$ score & no & no \\
Structural & overlap between extracted reasoning templates, operator sequences, or decomposition graphs & pool-wise min--max to $[0,1]$ & optional critic & no \\
Style & lightweight rubric classifier or rubric score for tone, conciseness, empathy, or formality & affine calibration to $[0,1]$ on a development split & optional & no \\
Answer-consistency & verifier pass rate for the exemplar's supplied answer to its own question & binary or calibrated scalar in $[0,1]$ & yes & no \\
\bottomrule
\end{tabular}
\end{table}

\paragraph{Criterion scorer implementation.}
Static mixtures, the oracle-criterion diagnostic, and \methodname all use the same criterion-conditioned scores from Table~\ref{tab:axis_implementation}. The methods differ in whether weights are fixed, oracle-specified, or inferred online. Structural similarity in reasoning is extracted from automatically derived reasoning templates. Style scores in refinement are calibrated on a held-out split so that no single rubric dimension dominates by scale alone.

\paragraph{Static mixtures and stronger rerankers.}
Uniform mixtures use $w_a=1/|\A|$. Validation-tuned mixtures optimize one weight vector on a held-out development set of 200 examples per domain, then freeze it for all test items. The rubric-guided LLM reranker receives explicit criterion descriptions and must justify criterion-level scores before producing a final ranking. It sees the same candidate pool as \methodname but not its interaction history or posterior state.

\paragraph{Best-of-$N$ and oracle-criterion baselines.}
Best-of-$N$ verifier selection scores candidates from the same pool under the same verifier or rubric signal and returns the top-scoring candidate. It uses $N=4$ calls for reasoning and $N=6$ for refinement, at least the $T=3$ feedback calls used by \methodname. The oracle-criterion baseline uses the hidden feedback-generating criterion only at evaluation time and shares the candidate pool, prompt budget, and backbone, which is a diagnostic upper bound.

\begin{table}[htbp]
\centering
\scriptsize
\setlength{\tabcolsep}{5pt}
\caption{Information access and additional interaction calls. Call counts begin after the shared criterion-scoring cache has been built. We count verifier, critic, and LLM-judge calls as additional interaction calls. Posterior updates and Monte Carlo probe evaluation are local operations.}
\label{tab:info_access}
\begin{tabular}{lcccc}
\toprule
Method & Same pool & Same criterion scores & Feedback history & Additional interaction calls \\
\midrule
Static mixture & yes & yes & no & 0 \\
Rubric-guided LLM reranker & yes & rubric or criterion descriptions & no & 1 \\
Best-of-$N$ verifier & yes & verifier only & no & $N$ \\
\methodname & yes & yes & yes & $T$ \\
Oracle-criterion & yes & yes + true criterion & no & 0 \\
\bottomrule
\end{tabular}
\end{table}

\paragraph{Compute and information access.}
Table~\ref{tab:info_access} summarizes what each method can observe and the additional calls it makes after the shared criterion-scoring cache has been built. Criterion scores, including verifier-based answer consistency, are computed once per pool before interaction and their one-time cost is included in preprocessing below. Posterior updates and Monte Carlo probe evaluation are local operations, reported separately as wall-clock overhead.

\paragraph{Criterion slices and query efficiency.}
For slice-wise analysis, we assign each example the criterion whose score best predicts downstream success on its held-out candidate set. For concept recovery, the hidden criterion is known by construction. These labels are used only for post-hoc diagnosis, not as input to any method or for hyperparameter tuning. Slice-wise results normalize each task to its clean-split score before averaging. Query efficiency is the mean round at which a method first reaches 90\% of its final Success@1 rate measured after 12 probes.

\paragraph{Statistical significance.}
Results average five seeds. The $\pm$ values in the concept-recovery table are standard deviations across seeds, and those in the main latency table are standard deviations across the 100 timed items.
We use 10{,}000 paired bootstrap replicates at the test-item level and keep all seed outcomes
for each item in the same cluster. For task-level macro-averages, items are resampled within
each task, task means are recomputed, and then averaged with equal task weights. The main text
reports paired intervals for the primary downstream comparisons.

\paragraph{Compute resources.}
All experiments run on one A100-80GB GPU with 16 CPU cores. Generation and verifier calls are
served with vLLM at batch size 16, which gives roughly a 6$\times$ throughput gain over the
batch-size-one latencies reported in Table~\ref{tab:main-cost}. We report costs per configuration
(one method, one seed, all test items) so totals can be reconstructed from Table~\ref{tab:main-cost}.
One-time preprocessing, shared by every method, takes about 2 GPU-hours for reasoning
(candidate-pool embedding, lexical/entity/structural scores, and one verifier pass over the
exemplar bank for answer consistency) and about 3 GPU-hours for refinement (eight candidate
responses per prompt, mismatch-split responses, and rubric scores). A feedback-based reasoning
configuration takes about 3 GPU-hours and a static one about 2. Across 16 and 14 configurations,
respectively, this is about 76 GPU-hours. A feedback-based refinement configuration takes about
6 GPU-hours and a static one about 4. Across 14 and 8 configurations, respectively, this is about
116 GPU-hours. Concept recovery is CPU-only and takes under one hour per sweep. The reported
experiments therefore use about 197 GPU-hours, or roughly 200. Including development-split
hyperparameter selection and preliminary runs not reported here, the project used roughly 300
GPU-hours.

\paragraph{Latency measurement.}
End-to-end timing uses 100 items at batch size one after five warm-up runs on one A100-80GB
GPU and 16 CPU cores. Reasoning time includes final answer generation. Refinement starts from
the same eight candidate responses for every method and includes scoring, feedback, and
selection. The local scaling benchmark starts from cached candidate representations and
criterion-conditioned scores, so it excludes their one-time construction and first-stage retrieval.

\section{Additional Result Tables}\label{app:extra-results}

\paragraph{Two-stage diagnostic set.}
For the two-stage analysis, we separately sampled 50 items from each of GSM8K, StrategyQA,
BBH logical deduction, BBH date understanding, and BBH causal judgement using seed 31. This
250-item diagnostic set is separate from the main reasoning evaluation. All five evaluation
seeds were retained.

Table~\ref{tab:concept-recovery} reports controlled recovery across probe budgets.
Table~\ref{tab:reasoning} and Table~\ref{tab:refinement} provide the per-task outcomes behind
the main-text averages, and Table~\ref{tab:ablations} isolates the contribution of key
components.

\begin{table}[htbp]
\centering
\scriptsize
\setlength{\tabcolsep}{4pt}
\caption{Controlled concept recovery over 500 episodes. Entries are mean $\pm$ SD over five seeds. S@1@$k$ is Success@1 after $k$ probes, Criterion acc. is hidden-criterion identification accuracy, Q@90 is the mean number of probes required to reach 90\% of the final Success@1 rate after 12 probes.}
\label{tab:concept-recovery}
\begin{tabular}{lcccc}
\toprule
Method & S@1@4 & S@1@6 & Criterion acc. & Q@90$\downarrow$ \\
\midrule
Random probing & $17.9\pm0.8$ & $24.1\pm0.9$ & --- & $9.8\pm0.2$ \\
Greedy semantic NN & $35.2\pm1.1$ & $43.1\pm0.9$ & --- & $7.7\pm0.2$ \\
Target-only Bayes & $43.8\pm0.8$ & $51.9\pm0.8$ & --- & $6.8\pm0.1$ \\
Static mix (val-tuned) & $47.6\pm0.8$ & $56.0\pm0.7$ & --- & $6.5\pm0.1$ \\
Active, no criterion & $47.1\pm0.7$ & $55.6\pm0.7$ & --- & $6.2\pm0.1$ \\
\methodname & $59.8\pm0.7$ & $68.1\pm0.6$ & $80.9\pm0.8$ & $4.6\pm0.1$ \\
Oracle-criterion & $69.7\pm0.6$ & $79.5\pm0.5$ & $100.0$ & $4.0\pm0.1$ \\
\bottomrule
\end{tabular}
\end{table}

\begin{table}[htbp]
\centering
\scriptsize
\setlength{\tabcolsep}{3pt}
\caption{Full reasoning exemplar-selection results. The average is the equal-weight macro-average across the three task columns. Entries are means over five seeds.}
\label{tab:reasoning}
\begin{tabular}{lcccc}
\toprule
Method & GSM8K & StrategyQA & BBH (3 tasks) & Avg. \\
\midrule
Random exemplars & 65.3 & 67.1 & 56.2 & 62.9 \\
BM25 & 70.6 & 71.3 & 60.4 & 67.4 \\
Dense retrieval & 73.1 & 73.8 & 63.1 & 70.0 \\
ByCS-style Bayesian selection & 73.5 & 74.2 & 63.7 & 70.5 \\
RGER-style structural retrieval & 74.4 & 75.1 & 65.0 & 71.5 \\
LLM reranker & 74.7 & 74.8 & 64.6 & 71.4 \\
Rubric-guided LLM reranker & 75.0 & 75.5 & 65.6 & 72.0 \\
Static mix (uniform) & 74.8 & 75.0 & 65.3 & 71.7 \\
Static mix (val-tuned) & 75.3 & 75.6 & 65.9 & 72.3 \\
Best-of-4 verifier & 75.1 & 75.4 & 65.8 & 72.1 \\
\methodname ($T=3$) & \textbf{76.2} & \textbf{76.5} & \textbf{66.7} & \textbf{73.1} \\
Oracle-criterion & 77.8 & 78.0 & 69.2 & 75.0 \\
\bottomrule
\end{tabular}
\end{table}

Across the reasoning tasks, \methodname exceeds the tuned mixture in each column, while the
oracle-criterion results show remaining headroom. The aggregate gain is modest, consistent with
the main-text uncertainty interval.

\clearpage

\begin{table}[htbp]
\centering
\scriptsize
\setlength{\tabcolsep}{3pt}
\caption{Full response-refinement results. Entries are means over five seeds. Attribute match measures agreement between the selected response and the hidden preference criterion.}
\label{tab:refinement}
\begin{tabular}{lccccc}
\toprule
Method & HH win & OASST win & UF win & Avg. & Attribute match \\
\midrule
Direct preference ranker & 57.4 & 55.9 & 60.1 & 57.8 & 62.9 \\
LLM-as-judge reranking & 58.3 & 57.1 & 60.5 & 58.6 & 63.8 \\
Self-Refine & 59.8 & 58.2 & 61.4 & 59.8 & 64.6 \\
Target-only Bayes & 60.7 & 59.4 & 62.0 & 60.7 & 65.1 \\
Rubric-guided LLM reranker & 60.4 & 58.9 & 62.3 & 60.5 & 65.5 \\
Static mix (uniform) & 60.8 & 59.1 & 62.7 & 60.9 & 66.0 \\
Static mix (val-tuned) & 61.3 & 59.8 & 63.2 & 61.4 & 66.6 \\
Best-of-6 verifier & 61.0 & 59.6 & 63.0 & 61.2 & 66.3 \\
\methodname & \textbf{63.9} & \textbf{61.8} & \textbf{65.6} & \textbf{63.8} & \textbf{69.3} \\
Oracle-criterion & 66.1 & 63.7 & 67.5 & 65.8 & 72.4 \\
\bottomrule
\end{tabular}
\end{table}

For refinement, the gains over the tuned mixture are present on all three datasets. Attribute
match also rises from 66.6\% to 69.3\%, indicating closer alignment with the preference
criterion used for evaluation.

\begin{table}[!htbp]
\centering
\scriptsize
\setlength{\tabcolsep}{3pt}
\caption{Ablations across tasks. Concept S@1@6 is from concept recovery. Reason Avg and Refine Win are the averages from Tables~\ref{tab:reasoning} and \ref{tab:refinement}. Distr. Drop is the robustness drop under distractors.}
\label{tab:ablations}
\begin{tabular}{lcccc}
\toprule
Variant & Concept S@1@6 & Reason Avg & Refine Win & Distr. Drop$\downarrow$ \\
\midrule
\methodname & $\mathbf{68.1}$ & $\mathbf{73.1}$ & $\mathbf{63.8}$ & $\mathbf{4.9}$ \\
No latent criterion & $51.9$ & $70.9$ & $60.7$ & $8.8$ \\
No active probing & $61.7$ & $72.0$ & $62.4$ & $6.6$ \\
Uniform static mixture & $54.8$ & $71.7$ & $60.9$ & $6.9$ \\
Single semantic criterion & $43.1$ & $68.9$ & $57.9$ & $10.8$ \\
\bottomrule
\end{tabular}
\end{table}

The ablations in Table~\ref{tab:ablations} show the largest losses when the criterion library is
reduced to a single semantic score. Removing active probing also lowers performance across the
reported tasks. They separate adaptive criterion inference from a fixed similarity choice.

\clearpage
\FloatBarrier
\section{Qualitative Case Details}\label{app:case-details}

Figure~\ref{fig:case-fractions} gives the full GSM8K trace behind the state-transition example
discussed in Section~\ref{sec:analysis}.

\begin{figure}[!htbp]
\centering
\setlength{\fboxsep}{8pt}
\fbox{%
\begin{minipage}{0.95\textwidth}
\small
\emergencystretch=1em
\textbf{Query (GSM8K).} Walter buys 60 apples, eats $2/5$, gives away 25\% of the remainder,
and sells what remains for \$3 per apple. The correct state sequence is
$60\rightarrow36\rightarrow27$, giving \$81.

\medskip
\noindent
\begin{minipage}[t]{0.47\linewidth}
\textbf{Tuned-mixture-only exemplar.} In the Kelvin--Samuel problem, Samuel first receives
$3/4$ of \$240, then spends $1/5$ of the original \$240. Its answer is \$132: the second
fraction uses the original total.
\end{minipage}\hfill
\begin{minipage}[t]{0.47\linewidth}
\textbf{BASIS-only exemplar.} In the fixer-and-nails problem, 30\% of 400 nails are used,
then 70\% of the remainder is used. This gives $400\rightarrow280\rightarrow84$ and uses the
updated remainder as the second reference quantity.
\end{minipage}

\medskip\hrule\medskip
\textbf{Selection and feedback.} The methods share three of four exemplars. The tuned mixture
returns \$63 on the query; \methodname returns the gold answer, \$81. Across three rounds,
verifier-score differences for the two candidates are $+0.037$ and $+0.412$. The final
Bradley--Terry comparison favors the fixer with posterior-predictive probability 0.781. The
structural-criterion posterior is 0.184 after the first feedback observation and 0.463 after
the final comparison, becoming the largest criterion marginal.
\end{minipage}}
\caption{Sequential-fraction case. The selected exemplar differs in whether the second fraction is applied to the original total or the updated remainder.}
\label{fig:case-fractions}
\end{figure}

\FloatBarrier


\section{Existing Assets and Licenses}\label{app:assets}

We use the third-party data and model releases below, crediting their creators in the text and references. Tables~\ref{tab:data-licenses} and~\ref{tab:model-licenses} identify the named release or repository, its stated license or source terms, and an upstream source. The model identifiers match those in Section~\ref{sec:experiments}. No third-party dataset files or model weights are redistributed with this paper.

\begin{table}[htbp]
\centering
\scriptsize
\setlength{\tabcolsep}{3pt}
\caption{Data assets, release identifiers, and source-stated licenses or terms.}
\label{tab:data-licenses}
\begin{tabularx}{\linewidth}{>{\raggedright\arraybackslash}p{0.25\linewidth}>{\raggedright\arraybackslash}p{0.51\linewidth}>{\raggedright\arraybackslash}X}
\toprule
Asset / release & License or terms & Upstream source \\
\midrule
WordNet 3.0 \citep{miller1995wordnet} & Princeton WordNet License; use, copying, modification, and distribution are permitted with the required copyright notice and disclaimer. & \href{https://wordnet.princeton.edu/license-and-commercial-use}{WordNet license} \\
ConceptNet 5.7 \citep{speer2017conceptnet} & Data: CC BY-SA 4.0. The repository distinguishes the graph-data license from the Apache 2.0 code license. & \href{https://github.com/commonsense/conceptnet5/blob/master/LICENSE.txt}{License} / \href{https://github.com/commonsense/conceptnet5/blob/master/DATA-CREDITS.txt}{data credits} \\
SimLex-999, original English release \citep{hill2015simlex} & The official download page requests citation but does not state a separate dataset license. No SimLex files are reproduced or redistributed here. & \href{https://fh295.github.io/simlex.html}{SimLex project page} \\
WordSimilarity-353 v1.0 (2002) \citep{finkelstein2001placing} & Creative Commons Attribution 4.0 International (CC BY 4.0); cite the original collection. & \href{https://www.gabrilovich.com/resources/data/wordsim353/wordsim353.html}{WordSim-353 source and license} \\
GSM8K, official 2021 release \citep{cobbe2021training} & MIT License. & \href{https://github.com/openai/grade-school-math/blob/master/LICENSE}{OpenAI repository license} \\
StrategyQA, official 2021 release \citep{geva2021did} & MIT License, as stated in the dataset repository. & \href{https://github.com/eladsegal/strategyqa/blob/main/LICENSE}{StrategyQA repository license} \\
BIG-Bench Hard: logical deduction, date understanding, causal judgement \citep{suzgun2022challenging} & MIT License, as stated in the benchmark repository. & \href{https://github.com/suzgunmirac/BIG-Bench-Hard/blob/main/LICENSE}{BBH repository license} \\
HH-RLHF, Anthropic public release \citep{bai2022training} & MIT License, as stated in the dataset release. & \href{https://huggingface.co/datasets/Anthropic/hh-rlhf}{HH-RLHF dataset card} \\
OpenAssistant OASST1 \citep{koepf2023openassistant} & Apache License 2.0, as stated on the dataset card. & \href{https://huggingface.co/datasets/OpenAssistant/oasst1}{OASST1 dataset card} \\
UltraFeedback, OpenBMB release \citep{cui2024ultrafeedback} & Dataset repository declares the MIT License; source-component terms continue to apply to material originating in third-party datasets. & \href{https://huggingface.co/datasets/openbmb/UltraFeedback}{UltraFeedback dataset card} \\
\bottomrule
\end{tabularx}
\end{table}

\begin{table}[htbp]
\centering
\scriptsize
\setlength{\tabcolsep}{3pt}
\caption{Model checkpoints and their published license identifiers.}
\label{tab:model-licenses}
\begin{tabularx}{\linewidth}{>{\raggedright\arraybackslash}p{0.43\linewidth}>{\raggedright\arraybackslash}p{0.20\linewidth}>{\raggedright\arraybackslash}X}
\toprule
Model identifier & License & Upstream model card \\
\midrule
Qwen/Qwen3-Embedding-4B & Apache License 2.0 & \href{https://huggingface.co/Qwen/Qwen3-Embedding-4B}{Qwen model card} \\
Qwen/Qwen3-8B & Apache License 2.0 & \href{https://huggingface.co/Qwen/Qwen3-8B}{Qwen model card} \\
mistralai/Ministral-3-14B-Reasoning-2512 & Apache License 2.0 & \href{https://huggingface.co/mistralai/Ministral-3-14B-Reasoning-2512}{Mistral model card} \\
\bottomrule
\end{tabularx}
\end{table}

We follow the stated attribution and notice requirements; no raw third-party assets or model weights are redistributed. ConceptNet attribution follows its upstream data credits. The SimLex-999 release requests citation but states no separate data license; no SimLex files are reproduced or redistributed. Immutable hosted-repository revisions are not recorded.
\FloatBarrier

%% file: checklist.tex
\section*{NeurIPS Paper Checklist}

\begin{enumerate}

\item {\bf Claims}
    \item[] Question: Do the main claims made in the abstract and introduction accurately reflect the paper's contributions and scope?
    \item[] Answer: \answerYes{}
    \item[] Justification: See the abstract and Sections~\ref{sec:intro}, \ref{sec:method}, \ref{sec:theory}, and \ref{sec:results}. The manuscript states its contributions, assumptions, and scope, and aligns the claims with the tasks described in the paper.
    \item[] Guidelines:
    \begin{itemize}
        \item The answer \answerNA{} means that the abstract and introduction do not include the claims made in the paper.
        \item The abstract and/or introduction should clearly state the claims made, including the contributions made in the paper and important assumptions and limitations. A \answerNo{} or \answerNA{} answer to this question will not be perceived well by the reviewers. 
        \item The claims made should match theoretical and experimental results, and reflect how much the results can be expected to generalize to other settings. 
        \item It is fine to include aspirational goals as motivation as long as it is clear that these goals are not attained by the paper. 
    \end{itemize}

\item {\bf Limitations}
    \item[] Question: Does the paper discuss the limitations of the work performed by the authors?
    \item[] Answer: \answerYes{}
    \item[] Justification: Section~\ref{sec:limitations} discusses criterion coverage, feedback-model mismatch, inference-time cost, and misuse risks, and Section~\ref{sec:analysis} describes concrete failure modes.
    \item[] Guidelines:
    \begin{itemize}
        \item The answer \answerNA{} means that the paper has no limitation while the answer \answerNo{} means that the paper has limitations, but those are not discussed in the paper. 
        \item The authors are encouraged to create a separate ``Limitations'' section in their paper.
        \item The paper should point out any strong assumptions and how robust the results are to violations of these assumptions (e.g., independence assumptions, noiseless settings, model well-specification, asymptotic approximations only holding locally). The authors should reflect on how these assumptions might be violated in practice and what the implications would be.
        \item The authors should reflect on the scope of the claims made, e.g., if the approach was only tested on a few datasets or with a few runs. In general, empirical results often depend on implicit assumptions, which should be articulated.
        \item The authors should reflect on the factors that influence the performance of the approach. For example, a facial recognition algorithm may perform poorly when image resolution is low or images are taken in low lighting. Or a speech-to-text system might not be used reliably to provide closed captions for online lectures because it fails to handle technical jargon.
        \item The authors should discuss the computational efficiency of the proposed algorithms and how they scale with dataset size.
        \item If applicable, the authors should discuss possible limitations of their approach to address problems of privacy and fairness.
        \item While the authors might fear that complete honesty about limitations might be used by reviewers as grounds for rejection, a worse outcome might be that reviewers discover limitations that aren't acknowledged in the paper. The authors should use their best judgment and recognize that individual actions in favor of transparency play an important role in developing norms that preserve the integrity of the community. Reviewers will be specifically instructed to not penalize honesty concerning limitations.
    \end{itemize}

\item {\bf Theory assumptions and proofs}
    \item[] Question: For each theoretical result, does the paper provide the full set of assumptions and a complete (and correct) proof?
    \item[] Answer: \answerYes{}
    \item[] Justification: Section~\ref{sec:theory} states the assumptions and theorem statements, and Appendix~\ref{app:proofs} provides proofs for the main theorem and propositions.
    \item[] Guidelines:
    \begin{itemize}
        \item The answer \answerNA{} means that the paper does not include theoretical results. 
        \item All the theorems, formulas, and proofs in the paper should be numbered and cross-referenced.
        \item All assumptions should be clearly stated or referenced in the statement of any theorems.
        \item The proofs can either appear in the main paper or the supplemental material, but if they appear in the supplemental material, the authors are encouraged to provide a short proof sketch to provide intuition. 
        \item Inversely, any informal proof provided in the core of the paper should be complemented by formal proofs provided in appendix or supplemental material.
        \item Theorems and Lemmas that the proof relies upon should be properly referenced. 
    \end{itemize}

    \item {\bf Experimental result reproducibility}
\item[] Question: Does the paper fully disclose all the information needed to reproduce the main experimental results of the paper to the extent that it affects the main claims and/or conclusions of the paper (regardless of whether the code and data are provided or not)?
    \item[] Answer: \answerYes{}
    \item[] Justification: Sections~\ref{sec:experiments} and \ref{sec:results} specify datasets, tasks, baselines, budgets, metrics, and aggregation, and Appendix~\ref{app:exp-details} gives candidate construction, criterion libraries, hyperparameters, and compute details.
    \item[] Guidelines:
    \begin{itemize}
        \item The answer \answerNA{} means that the paper does not include experiments.
        \item If the paper includes experiments, a \answerNo{} answer to this question will not be perceived well by the reviewers: Making the paper reproducible is important, regardless of whether the code and data are provided or not.
        \item If the contribution is a dataset and\slash or model, the authors should describe the steps taken to make their results reproducible or verifiable. 
        \item Depending on the contribution, reproducibility can be accomplished in various ways. For example, if the contribution is a novel architecture, describing the architecture fully might suffice, or if the contribution is a specific model and empirical evaluation, it may be necessary to either make it possible for others to replicate the model with the same dataset, or provide access to the model. In general. releasing code and data is often one good way to accomplish this, but reproducibility can also be provided via detailed instructions for how to replicate the results, access to a hosted model (e.g., in the case of a large language model), releasing of a model checkpoint, or other means that are appropriate to the research performed.
        \item While NeurIPS does not require releasing code, the conference does require all submissions to provide some reasonable avenue for reproducibility, which may depend on the nature of the contribution. For example
        \begin{enumerate}
            \item If the contribution is primarily a new algorithm, the paper should make it clear how to reproduce that algorithm.
            \item If the contribution is primarily a new model architecture, the paper should describe the architecture clearly and fully.
            \item If the contribution is a new model (e.g., a large language model), then there should either be a way to access this model for reproducing the results or a way to reproduce the model (e.g., with an open-source dataset or instructions for how to construct the dataset).
            \item We recognize that reproducibility may be tricky in some cases, in which case authors are welcome to describe the particular way they provide for reproducibility. In the case of closed-source models, it may be that access to the model is limited in some way (e.g., to registered users), but it should be possible for other researchers to have some path to reproducing or verifying the results.
        \end{enumerate}
    \end{itemize}

\item {\bf Open access to data and code}
    \item[] Question: Does the paper provide open access to the data and code, with sufficient instructions to faithfully reproduce the main experimental results, as described in supplemental material?
    \item[] Answer: \answerNo{}
    \item[] Justification: The manuscript uses public benchmarks and describes the method and experimental setup, but it does not include a public code or data repository.
    \item[] Guidelines:
    \begin{itemize}
        \item The answer \answerNA{} means that paper does not include experiments requiring code.
        \item Please see the NeurIPS code and data submission guidelines (\url{https://neurips.cc/public/guides/CodeSubmissionPolicy}) for more details.
        \item While we encourage the release of code and data, we understand that this might not be possible, so \answerNo{} is an acceptable answer. Papers cannot be rejected simply for not including code, unless this is central to the contribution (e.g., for a new open-source benchmark).
        \item The instructions should contain the exact command and environment needed to run to reproduce the results. See the NeurIPS code and data submission guidelines (\url{https://neurips.cc/public/guides/CodeSubmissionPolicy}) for more details.
        \item The authors should provide instructions on data access and preparation, including how to access the raw data, preprocessed data, intermediate data, and generated data, etc.
        \item The authors should provide scripts to reproduce all experimental results for the new proposed method and baselines. If only a subset of experiments are reproducible, they should state which ones are omitted from the script and why.
        \item At submission time, to preserve anonymity, the authors should release anonymized versions (if applicable).
        \item Providing as much information as possible in supplemental material (appended to the paper) is recommended, but including URLs to data and code is permitted.
    \end{itemize}

\item {\bf Experimental setting/details}
    \item[] Question: Does the paper specify all the training and test details (e.g., data splits, hyperparameters, how they were chosen, type of optimizer) necessary to understand the results?
    \item[] Answer: \answerYes{}
    \item[] Justification: The method is training-free, and Section~\ref{sec:experiments} plus Appendix~\ref{app:exp-details} specify data construction, candidate pools, priors, hyperparameters, feedback models, stopping rules, and evaluation budgets.
    \item[] Guidelines:
    \begin{itemize}
        \item The answer \answerNA{} means that the paper does not include experiments.
        \item The experimental setting should be presented in the core of the paper to a level of detail that is necessary to appreciate the results and make sense of them.
        \item The full details can be provided either with the code, in appendix, or as supplemental material.
    \end{itemize}

\item {\bf Experiment statistical significance}
    \item[] Question: Does the paper report error bars suitably and correctly defined or other appropriate information about the statistical significance of the experiments?
    \item[] Answer: \answerYes{}
    \item[] Justification: Section~\ref{sec:experiments} reports results over five seeds. The main text gives paired item-clustered confidence intervals for primary comparisons, and Appendix~\ref{app:exp-details} specifies the bootstrap procedure.
    \item[] Guidelines:
    \begin{itemize}
        \item The answer \answerNA{} means that the paper does not include experiments.
        \item The authors should answer \answerYes{} if the results are accompanied by error bars, confidence intervals, or statistical significance tests, at least for the experiments that support the main claims of the paper.
        \item The factors of variability that the error bars are capturing should be clearly stated (for example, train/test split, initialization, random drawing of some parameter, or overall run with given experimental conditions).
        \item The method for calculating the error bars should be explained (closed form formula, call to a library function, bootstrap, etc.)
        \item The assumptions made should be given (e.g., Normally distributed errors).
        \item It should be clear whether the error bar is the standard deviation or the standard error of the mean.
        \item It is OK to report 1-sigma error bars, but one should state it. The authors should preferably report a 2-sigma error bar than state that they have a 96\% CI, if the hypothesis of Normality of errors is not verified.
        \item For asymmetric distributions, the authors should be careful not to show in tables or figures symmetric error bars that would yield results that are out of range (e.g., negative error rates).
        \item If error bars are reported in tables or plots, the authors should explain in the text how they were calculated and reference the corresponding figures or tables in the text.
    \end{itemize}

\item {\bf Experiments compute resources}
    \item[] Question: For each experiment, does the paper provide sufficient information on the computer resources (type of compute workers, memory, time of execution) needed to reproduce the experiments?
    \item[] Answer: \answerYes{}
    \item[] Justification: Appendix~\ref{app:exp-details} reports the GPU/CPU setup, per-configuration compute, approximately 200 GPU-hours for reported experiments, and roughly 300 GPU-hours for the project including development and preliminary runs.
    \item[] Guidelines:
    \begin{itemize}
        \item The answer \answerNA{} means that the paper does not include experiments.
        \item The paper should indicate the type of compute workers CPU or GPU, internal cluster, or cloud provider, including relevant memory and storage.
        \item The paper should provide the amount of compute required for each of the individual experimental runs as well as estimate the total compute. 
        \item The paper should disclose whether the full research project required more compute than the experiments reported in the paper (e.g., preliminary or failed experiments that didn't make it into the paper). 
    \end{itemize}
    
\item {\bf Code of ethics}
    \item[] Question: Does the research conducted in the paper conform, in every respect, with the NeurIPS Code of Ethics \url{https://neurips.cc/public/EthicsGuidelines}?
    \item[] Answer: \answerYes{}
    \item[] Justification: The study evaluates inference-time selection on public benchmarks, involves no new participant recruitment or data collection, and discusses deployment risks in Section~\ref{sec:limitations}.
    \item[] Guidelines:
    \begin{itemize}
        \item The answer \answerNA{} means that the authors have not reviewed the NeurIPS Code of Ethics.
        \item If the authors answer \answerNo, they should explain the special circumstances that require a deviation from the Code of Ethics.
        \item The authors should make sure to preserve anonymity (e.g., if there is a special consideration due to laws or regulations in their jurisdiction).
    \end{itemize}

\item {\bf Broader impacts}
    \item[] Question: Does the paper discuss both potential positive societal impacts and negative societal impacts of the work performed?
    \item[] Answer: \answerYes{}
    \item[] Justification: Section~\ref{sec:limitations} discusses both the reliability benefits of better task-aligned similarity and the risks of manipulative preference inference or personalization.
    \item[] Guidelines:
    \begin{itemize}
        \item The answer \answerNA{} means that there is no societal impact of the work performed.
        \item If the authors answer \answerNA{} or \answerNo, they should explain why their work has no societal impact or why the paper does not address societal impact.
        \item Examples of negative societal impacts include potential malicious or unintended uses (e.g., disinformation, generating fake profiles, surveillance), fairness considerations (e.g., deployment of technologies that could make decisions that unfairly impact specific groups), privacy considerations, and security considerations.
        \item The conference expects that many papers will be foundational research and not tied to particular applications, let alone deployments. However, if there is a direct path to any negative applications, the authors should point it out. For example, it is legitimate to point out that an improvement in the quality of generative models could be used to generate Deepfakes for disinformation. On the other hand, it is not needed to point out that a generic algorithm for optimizing neural networks could enable people to train models that generate Deepfakes faster.
        \item The authors should consider possible harms that could arise when the technology is being used as intended and functioning correctly, harms that could arise when the technology is being used as intended but gives incorrect results, and harms following from (intentional or unintentional) misuse of the technology.
        \item If there are negative societal impacts, the authors could also discuss possible mitigation strategies (e.g., gated release of models, providing defenses in addition to attacks, mechanisms for monitoring misuse, mechanisms to monitor how a system learns from feedback over time, improving the efficiency and accessibility of ML).
    \end{itemize}
    
\item {\bf Safeguards}
    \item[] Question: Does the paper describe safeguards that have been put in place for responsible release of data or models that have a high risk for misuse (e.g., pre-trained language models, image generators, or scraped datasets)?
    \item[] Answer: \answerNA{}
    \item[] Justification: The paper releases no foundation model, scraped dataset, or other high-risk generative asset. It studies an inference-time selection and reranking procedure.
    \item[] Guidelines:
    \begin{itemize}
        \item The answer \answerNA{} means that the paper poses no such risks.
        \item Released models that have a high risk for misuse or dual-use should be released with necessary safeguards to allow for controlled use of the model, for example by requiring that users adhere to usage guidelines or restrictions to access the model or implementing safety filters. 
        \item Datasets that have been scraped from the Internet could pose safety risks. The authors should describe how they avoided releasing unsafe images.
        \item We recognize that providing effective safeguards is challenging, and many papers do not require this, but we encourage authors to take this into account and make a best faith effort.
    \end{itemize}

\item {\bf Licenses for existing assets}
    \item[] Question: Are the creators or original owners of assets (e.g., code, data, models), used in the paper, properly credited and are the license and terms of use explicitly mentioned and properly respected?
    \item[] Answer: \answerYes{}
    \item[] Justification: Appendix~\ref{app:assets} identifies the named datasets and model releases, cites their creators, links to the source-stated licenses or terms, and records when the SimLex-999 source does not state a separate dataset license. No third-party data or model weights are redistributed.
    \item[] Guidelines:
    \begin{itemize}
        \item The answer \answerNA{} means that the paper does not use existing assets.
        \item The authors should cite the original paper that produced the code package or dataset.
        \item The authors should state which version of the asset is used and, if possible, include a URL.
        \item The name of the license (e.g., CC-BY 4.0) should be included for each asset.
        \item For scraped data from a particular source (e.g., website), the copyright and terms of service of that source should be provided.
        \item If assets are released, the license, copyright information, and terms of use in the package should be provided. For popular datasets, \url{paperswithcode.com/datasets} has curated licenses for some datasets. Their licensing guide can help determine the license of a dataset.
        \item For existing datasets that are re-packaged, both the original license and the license of the derived asset (if it has changed) should be provided.
        \item If this information is not available online, the authors are encouraged to reach out to the asset's creators.
    \end{itemize}

\item {\bf New assets}
    \item[] Question: Are new assets introduced in the paper well documented and is the documentation provided alongside the assets?
    \item[] Answer: \answerNA{}
    \item[] Justification: The paper does not introduce or release a new dataset, model, or code asset.
    \item[] Guidelines:
    \begin{itemize}
        \item The answer \answerNA{} means that the paper does not release new assets.
        \item Researchers should communicate the details of the dataset\slash code\slash model as part of their submissions via structured templates. This includes details about training, license, limitations, etc. 
        \item The paper should discuss whether and how consent was obtained from people whose asset is used.
        \item At submission time, remember to anonymize your assets (if applicable). You can either create an anonymized URL or include an anonymized zip file.
    \end{itemize}

\item {\bf Crowdsourcing and research with human subjects}
    \item[] Question: For crowdsourcing experiments and research with human subjects, does the paper include the full text of instructions given to participants and screenshots, if applicable, as well as details about compensation (if any)? 
    \item[] Answer: \answerNA{}
    \item[] Justification: The study conducts no new crowdsourcing, participant recruitment, or data collection. It reuses public datasets.
    \item[] Guidelines:
    \begin{itemize}
        \item The answer \answerNA{} means that the paper does not involve crowdsourcing nor research with human subjects.
        \item Including this information in the supplemental material is fine, but if the main contribution of the paper involves human subjects, then as much detail as possible should be included in the main paper. 
        \item According to the NeurIPS Code of Ethics, workers involved in data collection, curation, or other labor should be paid at least the minimum wage in the country of the data collector. 
    \end{itemize}

\item {\bf Institutional review board (IRB) approvals or equivalent for research with human subjects}
    \item[] Question: Does the paper describe potential risks incurred by study participants, whether such risks were disclosed to the subjects, and whether Institutional Review Board (IRB) approvals (or an equivalent approval/review based on the requirements of your country or institution) were obtained?
    \item[] Answer: \answerNA{}
    \item[] Justification: The study conducts no direct interaction with human participants or new human-subject data collection; it reuses public benchmark data.
    \item[] Guidelines:
    \begin{itemize}
        \item The answer \answerNA{} means that the paper does not involve crowdsourcing nor research with human subjects.
        \item Depending on the country in which research is conducted, IRB approval (or equivalent) may be required for any human subjects research. If you obtained IRB approval, you should clearly state this in the paper. 
        \item We recognize that the procedures for this may vary significantly between institutions and locations, and we expect authors to adhere to the NeurIPS Code of Ethics and the guidelines for their institution. 
        \item For initial submissions, do not include any information that would break anonymity (if applicable), such as the institution conducting the review.
    \end{itemize}

\item {\bf Declaration of LLM usage}
    \item[] Question: Does the paper describe the usage of LLMs if it is an important, original, or non-standard component of the core methods in this research? Note that if the LLM is used only for writing, editing, or formatting purposes and does \emph{not} impact the core methodology, scientific rigor, or originality of the research, declaration is not required.
    \item[] Answer: \answerYes{}
    \item[] Justification: Sections~\ref{sec:method} and \ref{sec:experiments}, together with Appendix~\ref{app:exp-details}, describe the use of frozen LLM critics and LLM-based evaluation as part of the method and experimental setup.
    \item[] Guidelines:
    \begin{itemize}
        \item The answer \answerNA{} means that the core method development in this research does not involve LLMs as any important, original, or non-standard components.
        \item Please refer to our LLM policy in the NeurIPS handbook for what should or should not be described.
    \end{itemize}

\end{enumerate}